\pdfoutput=1

\documentclass[11pt]{article}

\usepackage{acl}

\usepackage{times}
\usepackage{latexsym}
\usepackage{amssymb}
\usepackage{booktabs}
\usepackage{colortbl}
\usepackage{array}
\usepackage{multirow}
\usepackage{graphicx}
\usepackage{dsfont}
\usepackage{paralist}
\usepackage{tikz}
\usetikzlibrary{arrows, arrows.meta, matrix, fit, positioning, decorations.pathreplacing, backgrounds, shapes, calc}
\newcolumntype{C}[1]{>{\centering\let\newline\\\arraybackslash\hspace{0pt}}m{#1}}
\newcolumntype{R}[1]{>{\raggedleft\let\newline\\\arraybackslash\hspace{0pt}}m{#1}}
\definecolor{mypink}{RGB}{252,225,225}
\definecolor{myblue}{RGB}{194,232,249}
\definecolor{myyellow}{RGB}{242,244,191}
\definecolor{mypurple}{RGB}{219,223,240}
\definecolor{myorange}{RGB}{254,227,184}

\usepackage[T1]{fontenc}

\usepackage[utf8]{inputenc}

\usepackage{microtype}

\usepackage{inconsolata}

\usepackage{amsmath}
\usepackage{csquotes}

\title{PETapter: Leveraging PET-style classification heads for modular few-shot parameter-efficient fine-tuning}

\author{Jonas Rieger \and Mattes Ruckdeschel \and Gregor Wiedemann \\
Leibniz Institute for Media Research | Hans-Bredow-Institut (HBI) \\
Rothenbaumchaussee 36, 20148 Hamburg, Germany\\
\texttt{rieger@statistik.tu-dortmund.de}\\
\texttt{\{m.ruckdeschel, g.wiedemann\} @leibniz-hbi.de}}

\begin{document}
\maketitle
\begin{abstract}
Few-shot learning and parameter-efficient fine-tuning (PEFT) are crucial to overcome the challenges of data scarcity and ever growing language model sizes. This applies in particular to specialized scientific domains, where researchers might lack expertise and resources to fine-tune high-performing language models to nuanced tasks. We propose PETapter, a novel method that effectively combines PEFT methods with PET-style classification heads to boost few-shot learning capabilities without the significant computational overhead typically associated with full model training. We validate our approach on three established NLP benchmark datasets and one real-world dataset from communication research. We show that PETapter not only achieves comparable performance to full few-shot fine-tuning using pattern-exploiting training (PET), but also provides greater reliability and higher parameter efficiency while enabling higher modularity and easy sharing of the trained modules, which enables more researchers to utilize high-performing NLP-methods in their research.
\end{abstract}

\section{Introduction}
Few-shot learning and parameter-efficient fine-tuning (PEFT) have emerged as pivotal disciplines in the realm of natural language processing (NLP), especially in applications where data scarcity poses significant challenges. Whilst for a lot of scientists from fields like communication science or political science, raw text data is available, high-quality labeled data is still scarce and the creation is time-consuming, costly and requiring trained experts. Discourse analysis in those fields, is closely related to the NLP sub-task of argument mining, especially argument identification and classification. Due to the nuanced and context-dependent nature of argumentative discourse, standard fine-tuning methods for pretrained language models (PLM) with only a few labeled data points usually perform poorly in these tasks \cite[e.g.,][]{rieger2023fewshot}. Few-shot learning, with its promise to generalize from a limited number of training samples, offers an alternative pathway towards mitigating the data scarcity problem. However, the deployment of large-scale language models in a few-shot setting can be challenging, primarily due to the substantial computational resources required for training and fine-tuning, as they update all parameters of the model by default. PEFT methods, such as adapter modules, present a viable solution to this issue by enabling the adaptation of pretrained models to specific tasks with minimal modifications to the model architecture. Specifically, these methods freeze a large part of the model's parameters and train only small parts of the existing or few newly added layers \cite{poth-etal-2023-adapters}. Further, PEFT methods using a standard linear layer perform significantly worse than few-shot methods when using few observations (cf.~Sections~\ref{sec:results_ag_yahoo_yelp} and~\ref{sec:results_ukraine}).

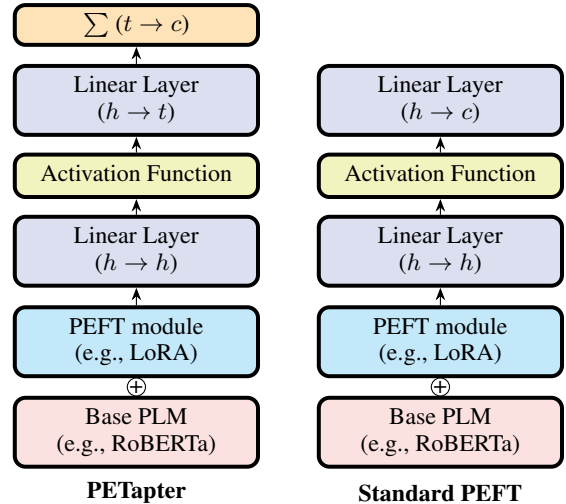
\begin{figure}[t]
\small
    \centering
    \begin{tikzpicture}
        \node (base) at (0,-0.8) {\bf PETapter};
        \node[draw, rounded corners, line width=1.5, fill=mypink, minimum width=32mm] (base) at (0,0) {\begin{tabular}{c} Base PLM \\(e.g., RoBERTa)\end{tabular}};
        \node[draw, rounded corners, line width=1.5, fill=myblue, minimum width=32mm] (peft) at (0,1.2) {\begin{tabular}{c} PEFT module \\(e.g., LoRA)\end{tabular}};
        \node[draw, rounded corners, line width=1.5, fill=mypurple, minimum width=32mm] (linear1) at (0,2.4) {\begin{tabular}{c} Linear Layer \\($h \to h$)\end{tabular}};
        \node[draw, rounded corners, line width=1.5, fill=myyellow, minimum width=32mm] (gelu) at (0,3.4) {\begin{tabular}{c} Activation Function\end{tabular}};
        \node[draw, rounded corners, line width=1.5, fill=mypurple, minimum width=32mm] (linear2) at (0,4.4) {\begin{tabular}{c} Linear Layer \\($h \to t$)\end{tabular}};
        \node[draw, rounded corners, line width=1.5, fill=myorange, minimum width=32mm] (mapping) at (0,5.4) {$\sum$ ($t \to c$)};
        \path (base) -- (peft) node[draw, circle, inner sep=0pt, minimum size=1pt, midway] {+};
        \draw[-Stealth] (peft) -- (linear1);
        \draw[-Stealth] (linear1) -- (gelu);
        \draw[-Stealth] (gelu) -- (linear2);
        \draw[-Stealth] (linear2) -- (mapping);

        \node (base) at (4,-0.8) {\bf Standard PEFT};
        \node[draw, rounded corners, line width=1.5, fill=mypink, minimum width=32mm] (base) at (4,0) {\begin{tabular}{c} Base PLM \\(e.g., RoBERTa)\end{tabular}};
        \node[draw, rounded corners, line width=1.5, fill=myblue, minimum width=32mm] (peft) at (4,1.2) {\begin{tabular}{c} PEFT module \\(e.g., LoRA)\end{tabular}};
        \node[draw, rounded corners, line width=1.5, fill=mypurple, minimum width=32mm] (linear1) at (4,2.4) {\begin{tabular}{c} Linear Layer \\($h \to h$)\end{tabular}};
        \node[draw, rounded corners, line width=1.5, fill=myyellow, minimum width=32mm] (tanh) at (4,3.4) {\begin{tabular}{c} Activation Function\end{tabular}};
        \node[draw, rounded corners, line width=1.5, fill=mypurple, minimum width=32mm] (linear2) at (4,4.4) {\begin{tabular}{c} Linear Layer \\($h \to c$)\end{tabular}};
        \path (base) -- (peft) node[draw, circle, inner sep=0pt, minimum size=1pt, midway] {+};
        \draw[-Stealth] (peft) -- (linear1);
        \draw[-Stealth] (linear1) -- (tanh);
        \draw[-Stealth] (tanh) -- (linear2);
    \end{tikzpicture}
    \caption{Schematic representation of the PETapter architecture compared to standard PEFT methods.}
    \label{fig:petapter}
\end{figure}

For this reason, this paper aims to delve into the innovative intersection of few-shot learning and parameter-efficient fine-tuning. By integrating few-shot learning paradigms with PEFT techniques, this paper explores how, e.g., the field of argument mining can leverage the strengths of both approaches to efficiently classify text sequences, even when little data is available. Through a study on three typical NLP benchmark datasets and a study on a real-world argument mining dataset, we demonstrate that the combination of few-shot and PEFT methodologies can significantly reduce the reliance on extensive annotated datasets while achieving competitive performance (to full few-shot fine-tuning) with reduced computational effort. It thus combines the advantages of the two fields, few-shot learning and parameter-efficient fine-tuning. This makes possible the creation of small yet powerful language models that can easily be created by non-experts, opening up new qualitative and quantitative research opportunities to researchers from various backgrounds who want to leverage text data.

As contribution, we propose a new method, PETapter, as a combination of PET-style classification heads with PEFT methods. We show that in most scenarios PETapter is as performant as PET itself, i.e.\ noticeable better than PEFT with a standard classification head. PETapter additionally offers all PEFT advantages, i.e.\ it requires less computational resources and is easier to share in the research community due to its modularity. We also show that PETapter provides more robust predictions than PET, especially on real-world datasets. By doing so, this paper not only contributes to the theoretical advancement of NLP techniques but also offers practical insights for researchers and practitioners working on supervised text classification problems from various scientific backgrounds.

In Section~\ref{sec:related_work}, we review relevant preliminary NLP work, in Section~\ref{sec:pvp}, we recapitulate the definition of pattern-verbalizer pairs (PVP) for usage in our new method PETapter in Section~\ref{sec:petapter}. In Sections~\ref{sec:lab} and~\ref{sec:ukraine}, we conduct intensive NLP benchmark studies and a real-world study, which we finally discuss in Section~\ref{sec:discussion} and summarize in Section~\ref{sec:conclusion}. The data and code used will be made available in a GitHub repository after acceptance.

\section{Related Work}
\label{sec:related_work}
Pretrained language models (PLMs) serve as the basis for most classification tasks in natural language processing (NLP). A frequently compared model is RoBERTa \cite{liu2019roberta} due to its strong performance even with a comparatively small number of parameters (Base: 125~million, Large: 355~million). When using non-English texts, the multilingual alternative XLM-RoBERTa \cite{conneau-etal-2020-unsupervised} can be used, which was trained on 100 different languages and provides a reliable basis for achieving good performance, at least for high resource languages. As the next best model, (m)DeBERTa \cite{he2023debertav3} forms a promising base PLM, but with 1.5~billion parameters it already has significantly higher requirements in terms of GPU capacity and computation time.

\subsection{Parameter-Efficient Fine-Tuning (PEFT)}
\label{sec:peft}
The ever-increasing sizes of base PLMs pose a great challenge for fully fine-tuning all parameters. \citet{rebuffi2017}, as well as \citet{houlsby2019}, were among the first to tackle this problem by freezing the PLM and instead inserting and training bottleneck feed-forward layers within the pretrained architecture. \citet{pfeiffer-etal-2020-adapterhub} refined this idea into a slightly more parameter-efficient and robust method, though still sequential in its basic idea, which is named Pfeiffer adapters. Techniques such as low-rank adaptation \cite[LoRA,][]{hu2021lora} and $(\text{IA})^3$ \cite{liu2022ia3} introduce minimal updates to the model weights, while their parallel approach makes it possible to combine the newly learned parameters with the frozen ones in such a way that there is no overhead during inference. Recent innovations such as considering different ranks for each linear layer in PRILoRA \cite{benedek-wolf-2024-prilora}, sharing the low-rank matrices across all layers with layer-specific learned scaling vectors in VeRA \cite{kopiczko2024vera} or LoRA-like decomposition of pretrained weights in DoRA \cite{liu2024dora} are examples of further LoRA-like advancements in the field of parameter-efficient fine-tuning.

In addition, ComPEFT \cite{yadav2023compeft} and QLoRA \cite{dettmers2023qlora} offer the possibility of even more effective parameter usage with condensation of information through quantization and desparsification. RoSA \cite{nikdan2024rosa} combines low-rank adaptation with high sparsity training following the idea of robust principal component analysis. Several specialized approaches exist, one of which is AdaSent \cite{huang-etal-2023-adasent} focusing on PEFT for sentence representations.

While many implementations are often limited to the PEFT character of the methods \cite{huggingfacepeft,hu-etal-2023-opendelta,hu-etal-2023-llm}, the Python package Adapters \cite{poth-etal-2023-adapters} facilitates the modular use of PEFT methods, highlighting a trend towards modular deep learning \cite{pfeiffer2023modular} and holistic PEFT implementations \cite{he2022towards,sabry2023peftref}.

\subsection{Few-Shot Text Classification}
\label{sec:few}
Few-shot text classification aims to maximize model performance with minimal labeled data. For this, a lot of approaches are based on a combination of prompt-based learning and meta learning. \citet{schick-schutze-2021-exploiting} are the first to propose the utilization of so-called \emph{prompt-based} cloze questions. In this pattern-exploiting training (PET), a masked token is predicted in a language model task style, similar to that used in pretraining. The authors show that due to the model's prior knowledge, it is able to predict these so-called verbalizers better than plain numbered class labels. Furthermore, the authors show that \emph{meta-learning} on soft labels generated by augmentations using iterative PET (iPET) also leads to an improvement, so that a combination of prompt-based learning and meta-learning \cite{zhang-etal-2022-prompt-based} is proposed. \citet{chen-shu-2023-promptda} use such an approach and propose the use of label-guided data augmentation methods for prompt-based few-shot tuning. 

\citet{ling2023} demonstrate the possibility of an optimized automated search for verbalizers in prompt-based learning, while \citet{karimi-mahabadi-etal-2022-prompt} suggest an approach without the need for prompts. Complementary to this, SetFit \cite{tunstall2022efficient} form triplets of positive and negative examples and use contrastive learning to effectively train on few training samples. The recipe T-Few \cite{liu2022ia3} as an additional dedicated few-shot model is based on the zero-shot model T0 \cite{sanh2022multitask} and enriches it with a combination of losses, the PEFT method $(\text{IA})^3$ and a pretraining of it using out-of-domain data, which can be quite expensive with regard to the training time needed. PET, Setfit and T-Few represent key developments in the few-shot discipline and produce comparable performances on the RAFT benchmark dataset \cite{alex2021raft}, each with strengths and weaknesses on different datasets

\subsection{Argument Mining}
\label{sec:arg}
In the field of argument mining, the following works are related to our idea of using PETapter to identify argumentative sentences from news media: \citet{huening2022detecting} classify whether user-generated chat messages contain an argument using machine learning techniques utilizing sentence embeddings as features. For simplicity, they categorize claims as \emph{no argument} in their setting. \citet{jurkschat-etal-2022-shot} consider the possibility of few-shot methods for classifying aspects of argumentative sentences and show that PET performs best among the methods considered. Likewise, PET is used by \citet{rieger2023fewshot} and compared with the use of PEFT methods for the identification of claims and arguments from news media articles. The authors find that the task is challenging even for humans, so that only moderate inter-coder agreements are achieved.

In the following, we present the PETapter model, which combines few-shot and PEFT methods to achieve great performance, in particular in real-world data scarce argument mining settings.

\section{Pattern-Verbalizer-Pair (PVP)}
\label{sec:pvp}
We consider a pretrained language model (PLM) $M$ with an underlying vocabulary $V$, where [MASK] $\in V$, i.e. we assume a mask token to be contained in the vocabulary. In addition, let $\mathcal{L}$ be a set of target labels for a classification task and $x \in V^n$ an input (possibly consisting of several segments) that contains a total of $n$ tokens from the vocabulary $V$. We define a pattern as a function $P^m: V^n \to V^{n+p}$, where $m \leq p$ is the number of [MASK] tokens contained in the pattern and $p$ is the total number of vocabulary items added to the input $x$ by the pattern function.

Furthermore, let $v^m:\mathcal{L} \to V^m$ be a verbalizer function that assigns $m$ tokens from the vocabulary $V$ to each label $\ell \in \mathcal{L}$. As the inventors of PET \cite{schick-schutze-2021-exploiting} suggest, we refer to the combination $(P^m, v^m)$ as a pattern-verbalizer pair (PVP). The pattern function $P^m$ is used to transform the input $x$ into a kind of cloze question and the verbalizer function $v^m$ provides for each label the \enquote{speaking} and representative fill-in words for the corresponding cloze question, which are to be predicted by the model $M$. All PVPs used in this paper can be found in the Appendices~\ref{sec:appendix_ag_yelp_yahoo} and~\ref{sec:appendix_ukraine}.

\section{PETapter}
\label{sec:petapter}
PETapter can be seen as a modular classification head which can be added flexibly to a PEFT framework as an alternative to a classical linear layer classification head, cf.~Figure~\ref{fig:petapter}. Here, the last (purple) linear layer of the classification head does not reduce the dimension directly to the number of classes $c$, but to the size $t$ of the sub-vocabulary $T$ of the verbalizer function used. Thus, each dimension of the output embedding represents a token from the reduced verbalizer vocabulary, similar to and motivated by the pattern-exploiting training (PET) by \citet{schick-schutze-2021-exploiting}. Then, the logits for the verbalizers (possibly composed of several tokens) can be calculated as a sum of the logits of the masked tokens in the pattern (cf.~Equation~\ref{eq:score}).

In general, PETapter can be combined with any PEFT method (cf.~the blue segment in Figure~\ref{fig:petapter}), i.e.~it also features all its advantages over full fine-tuning: faster training, less resource requirements, better sharability and reusability due to modularity as well as robustification (at least) with regard to the elimination of catastrophic forgetting. Specifically, PETapter implements all these advantages over PET, which uses full fine-tuning, while performing on par (cf.~Sections~\ref{sec:results_ag_yahoo_yelp} and~\ref{sec:results_ukraine}).

Let $\mathcal{L}$ be a set of target labels for a classification task with $\vert \mathcal{L} \vert = c$, let $P^m(x)$ be a pattern function inserting $m$ [MASK] tokens to an input $x$, and let $v^m(\ell)$ be an injective function that maps each of the labels $\ell \in \mathcal{L}$ to $m$ vocabulary tokens. Then, we obtain the subset of the vocabulary relevant for the verbalizers as $T = \bigcup_{\ell \in \mathcal{L}} \bigcup_{i=1}^m v^m(\ell)_i$ with $T \subset V$ and $t = \vert T \vert$ the corresponding number of relevant tokens, i.e.\ we can refine $v^m:\mathcal{L} \to T^m$.

Moreover, let $M(v^m(\ell) \mid P^m(x)) \in \mathbb{R}^m$ denote the logits for the $m$ [MASK] tokens, the output of the top linear layer (purple) in Figure~\ref{fig:petapter} on the left. The final score for each of the label candidates $\ell$ for a given input text $x$ is then given as
\begin{align}
\label{eq:score}
    s(\ell \mid x) &= \sum_{i=1}^m M(v^m(\ell) \mid P^m(x))_i
\end{align}
and the corresponding pseudo-probability as
\begin{align}
    q(\ell \mid x) &= \frac{\exp(s(\ell \mid x))}{\sum_{\ell'\in \mathcal{L}}\exp(s(\ell' \mid x))}.
\end{align}
Using this, we can calculate the cross-entropy loss over all observations as
\begin{align}
    L_\text{CE} &= -\sum_{(x,\ell^*)} \sum_{\ell \in \mathcal{L}} \mathds{1}_{\{\ell = \ell^*\}}(x,\ell^*) \log[ q(\ell \mid x) ] \notag \\
    &= -\sum_{(x,\ell^*)} \log[ q(\ell^* \mid x) ]
\end{align}
if we consider $\ell^* \in \mathcal{L}$ to be the true label of $x$.

\section{Benchmark Study}
\label{sec:lab}
To evaluate how well our model performs in comparison to existing state-of-the-art methods, we consider three established NLP datasets in a first benchmark study. These have quite a laboratory character, as the distribution of the class labels is pretty much balanced. Furthermore, it is not finally clear to what extent (possibly implicit) information about the test data of such publicly available datasets is contained in PLMs \cite{Li_Flanigan_2024}. Nevertheless, such datasets with predefined train-test splits offer the possibility of comparing methods across studies without the need of rerunning all the experiments.

\subsection{Datasets: AG, Yahoo, Yelp}
We follow the study by \citet{schick-schutze-2022-true} and use the three established NLP datasets \emph{AG's News} (AG), \emph{Yahoo Questions} (Yahoo), and \emph{Yelp Full} (Yelp), which are presented by \citet{datasets}. In Appendix~\ref{sec:appendix_ag_yelp_yahoo}, further information on the three datasets is given.

For the training, we create 5 stratified datasets with $n=10$ randomly drawn observations and 5 datasets with $n=100$ observations each for the problems AG, Yahoo, and Yelp. We evaluate all experiments with the corresponding entire test dataset from the predefined split.

\subsection{Experimental Setup}
\label{sec:lab_exp}
As PLM, we make use of RoBERTa Base and Large. For fine-tuning methods, we consider the few-shot method PET, PEFT methods with a (standard) linear layer as classification head, and our method PETapter, i.e.\ PEFT methods in combination with a PET-style classification head. In both cases, we consider the methods $(\text{IA})^3$, LoRA and a Pfeiffer adapter as PEFT methods. For PET and PETapter, we compare the use of \emph{prompt} or \emph{Q\&A} pattern, following \citet{schick-schutze-2022-true}. 
In each dataset, we measure the change of using $n=10$ or $100$ observations. Moreover, we repeat each experiment five times, i.e.\ in Table~\ref{tab:acc_clinic} each cell corresponds to 25 experiments (5 repetitions $\times$ 5 datasets). In Appendix~\ref{sec:appendix_ag_yelp_yahoo}, further information on all parameters as well as the used patterns is given.

\begin{table*}[t]
\setlength{\tabcolsep}{4pt}
    \centering
    \small
    \begin{tabular}{ccc|*{3}{R{2.6em}}|R{2.6em}|*{3}{R{2.6em}}|R{2.6em}|*{3}{R{2.6em}}}
    \toprule
  &&& \multicolumn{4}{c|}{\underline{Prompt Pattern}} & \multicolumn{4}{c|}{\underline{Q\&A Pattern}} & \multicolumn{3}{c}{\underline{Linear Layer}}\\
  &&& \multicolumn{3}{c}{PETapter} & & \multicolumn{3}{c}{PETapter} &&\\
  & $n$ & Data & \centering (IA)$^3$ & \centering LoRA & \centering Pfeif. & \centering PET & \centering (IA)$^3$ & \centering LoRA & \centering Pfeif. & \centering PET & \centering (IA)$^3$ & \centering LoRA & \multicolumn{1}{c}{Pfeif.}\\
  \hline
  \multirow{11}{*}{\rotatebox[origin=c]{90}{RoBERTa Base}} & 10 & AG & 0.668 $\pm$.092 & 0.681 $\pm$.080 & 0.683 $\pm$.078 & \bf 0.804 $\pm$.036 & 0.596 $\pm$.077 & 0.672 $\pm$.077 & 0.680 $\pm$.088 & 0.800 $\pm$.028 & 0.297 $\pm$.053 & 0.293 $\pm$.045 & 0.316 $\pm$.059 \\
  \arrayrulecolor{lightgray}
  \cline{4-14}
  & 10 & Yahoo & 0.236 $\pm$.016 & 0.292 $\pm$.033 & 0.271 $\pm$.042 & \bf 0.545 $\pm$.040 & 0.274 $\pm$.026 & 0.295 $\pm$.041 & 0.285 $\pm$.036 & 0.515 $\pm$.036 & 0.105 $\pm$.009 & 0.116 $\pm$.020 & 0.123 $\pm$.016 \\
  \cline{4-14}
  & 10 & Yelp & 0.403 $\pm$.037 & 0.434 $\pm$.035 & 0.423 $\pm$.038 & 0.441 $\pm$.017 & 0.359 $\pm$.037 & 0.412 $\pm$.042 & 0.390 $\pm$.047 & \bf 0.442 $\pm$.025 & 0.215 $\pm$.017 & 0.219 $\pm$.023 & 0.224 $\pm$.027 \\
  \cline{4-14}
  & 100 & AG & 0.856 $\pm$.012 & 0.861 $\pm$.012 & 0.860 $\pm$.012 & \bf 0.875 $\pm$.007 & 0.863 $\pm$.006 & 0.861 $\pm$.010 & 0.862 $\pm$.011 & 0.871 $\pm$.008 & 0.837 $\pm$.007 & 0.862 $\pm$.008 & 0.864 $\pm$.008 \\
  \cline{4-14}
  & 100 & Yahoo & 0.622 $\pm$.018 & 0.642 $\pm$.009 & 0.642 $\pm$.011 & \bf 0.666 $\pm$.007 & 0.622 $\pm$.005 & 0.637 $\pm$.007 & 0.633 $\pm$.008 & 0.659 $\pm$.007 & 0.418 $\pm$.034 & 0.636 $\pm$.013 & 0.633 $\pm$.012 \\
  \cline{4-14}
  & 100 & Yelp & 0.541 $\pm$.010 & 0.553 $\pm$.018 & 0.551 $\pm$.016 & 0.552 $\pm$.014 & 0.536 $\pm$.014 & 0.553 $\pm$.015 & 0.548 $\pm$.014 & \bf 0.554 $\pm$.015 & 0.354 $\pm$.022 & 0.538 $\pm$.018 & 0.535 $\pm$.020 \\
  \arrayrulecolor{black}
  \hline
  \multirow{11}{*}{\rotatebox[origin=c]{90}{RoBERTa Large}} & 10 & AG & 0.641 $\pm$.100 & 0.714 $\pm$.070 & 0.702 $\pm$.081 & \bf 0.842 $\pm$.025 & 0.611 $\pm$.073 & 0.746 $\pm$.054 & 0.738 $\pm$.060 & 0.836 $\pm$.032 & 0.305 $\pm$.030 & 0.373 $\pm$.049 & 0.443 $\pm$.104 \\
  \arrayrulecolor{lightgray}
  \cline{4-14}
  & 10 & Yahoo & 0.242 $\pm$.027 & 0.331 $\pm$.040 & 0.290 $\pm$.056 & \bf 0.574 $\pm$.030 & 0.323 $\pm$.049 & 0.365 $\pm$.049 & 0.346 $\pm$.054 & 0.550 $\pm$.040 & 0.124 $\pm$.012 & 0.150 $\pm$.027 & 0.169 $\pm$.041 \\
  \cline{4-14}
  & 10 & Yelp & 0.442 $\pm$.040 & 0.470 $\pm$.041 & 0.479 $\pm$.035 & 0.475 $\pm$.026 & 0.440 $\pm$.049 & 0.472 $\pm$.049 & \bf 0.490 $\pm$.046 & 0.486 $\pm$.041 & 0.211 $\pm$.010 & 0.221 $\pm$.012 & 0.216 $\pm$.014 \\
  \cline{4-14}
  & 100 & AG & 0.868 $\pm$.011 & 0.873 $\pm$.010 & 0.875 $\pm$.010 & \bf 0.877 $\pm$.009 & 0.876 $\pm$.009 & 0.870 $\pm$.010 & 0.873 $\pm$.010 & 0.874 $\pm$.009 & 0.833 $\pm$.011 & 0.875 $\pm$.008 & 0.875 $\pm$.008 \\
  \cline{4-14}
  & 100 & Yahoo & 0.654 $\pm$.020 & 0.662 $\pm$.014 & 0.661 $\pm$.017 & \bf 0.680 $\pm$.013 & 0.655 $\pm$.010 & 0.654 $\pm$.008 & 0.656 $\pm$.012 & 0.675 $\pm$.013 & 0.364 $\pm$.043 & 0.648 $\pm$.016 & 0.647 $\pm$.015 \\
  \cline{4-14}
  & 100 & Yelp & 0.611 $\pm$.011 & 0.613 $\pm$.014 & 0.614 $\pm$.010 & 0.593 $\pm$.014 & \bf 0.626 $\pm$.008 & 0.622 $\pm$.013 & 0.620 $\pm$.013 & 0.595 $\pm$.016 & 0.347 $\pm$.019 & 0.551 $\pm$.019 & 0.512 $\pm$.043 \\
  \arrayrulecolor{black}
  \bottomrule
    \end{tabular}
    \caption{Mean accuracies ($\pm$ standard deviation) of the experiments in the benchmark study.}
    \label{tab:acc_clinic}
\end{table*}

\begin{table}[t]
\small
    \centering
    \begin{tabular}{clccc}
    \toprule
    RoBERTa & Architecture & AG & Yahoo & Yelp \\
    \midrule
    Base & PETapter & 0.33 & 0.33 & 0.32\\
    Base & PET & 0.38 & 0.39 & 0.39\\
    Large & PETapter & 0.65 & 0.64 & 0.65\\
    Large & PET & 1.00 & 1.00 & 1.00\\
    \midrule
    \midrule
    Large & PET & 6.1s & 6.2s & 6.2s\\
    \bottomrule
    \end{tabular}
    \caption{Comparison of training times per iteration of $n=100$ observations in the benchmark study. The last row shows the time for PET using RoBERTa Large. The other rows indicate the time relative to it. The times for using PETapter or a linear layer are identical, as are for the three architectures $(\text{IA})^3$, LoRA, and Pfeiffer.}
    \label{tab:times_clinic}
\end{table}

\subsection{Results}
\label{sec:results_ag_yahoo_yelp}
Table~\ref{tab:acc_clinic} shows the average accuracy of the settings and the standard deviation across the 25 experiments per cell. Overall, it can be seen that the best performance is achieved by PET using the prompt pattern, while PETapter achieves the best values in two scenarios and the linear layer classification head never performs best. As expected, RoBERTa Large consistently yields better results than the Base variant, although the PEFT methods generally benefit somewhat more from the use of the larger model. The accuracy achieved by PETapter is usually very close to that of PET. Merely in the setting $n=10$ for Yahoo the values are significantly worse. On the other hand, for $n=100$ on Yelp, all three PEFT architectures combined with our PETapter method are noticeably better than PET itself. $(\text{IA})^3$ with Q\&A pattern even leads to the best global performance in this case, while it produces rather low scores in most scenarios. Our explanation for this is that IA3 is just one component of the generally in benchmarks competitively performing T-Few. Overall, prompt and Q\&A pattern perform similarly. Thus, we can confirm these findings from \citet{schick-schutze-2022-true} regarding PET and show that they generally hold for PETapter as well. Basically, it can be seen that the linear layer rarely comes close to the performance of PETapter; using linear layer classification heads in combination with $(\text{IA})^3$ leads to the worst results overall.

As a result, it is evident that PETapter is to be preferred over the use of a classical linear layer as a classification head in use cases where the benefits of PEFT methods are desired. At the same time, PETapter achieves comparable performance to PET in most cases.

In addition, as a PEFT method (cf.~Section~\ref{sec:peft}), PETapter can be trained more effectively than standard PET. In Table~\ref{tab:times_clinic} the training times of the models are displayed. According to this, PETapter (no matter the architecture) requires $\approx 65\%$ of the time of regular PET per iteration. Compared to a regular PEFT approach using a linear layer, PETapter does not produce any overhead in training time.

\section{Real-World Study}
\label{sec:ukraine}
As indicated in Section~\ref{sec:lab}, the evaluation based on AG, Yahoo, and Yelp represents a lab situation, e.g., because of their balancedness and (implicit) inclusion in the pretraining of language models. Complementary to this, in a second study, we compare the performance of the methods on a (so far unpublished) dataset with real-world challenges. It is a dataset in the context of argument mining with argumentative sentences on the topic of \emph{arms deliveries to Ukraine} with a total of 7301 thematically relevant articles in 2022 from 22 German media outlets. The composition is explained in detail in the study of \citet{rieger2023fewshot}. Here, we consider a version of the dataset consisting of 1766 labeled data with the four possible labels claim/argument for/against, with non-relevant sentences already removed.

\subsection{Dataset: Ukraine Arms Deliveries}
The German language Ukraine dataset consists of 766 train (294 articles) and 1000 test (369 articles) observations. Using a two-stage sampling (first article, then sentence) we intend to tackle the problem that including very related sentences from the same text in both splits could lead to an overoptimistic estimation of the error. Here, a single observation consists of a target sentence to be classified as well as two contextual sentences before and after it. Based on the 766 observations, we draw training sets of sizes $n=10, 100, 250$ according to three different sampling strategies.

We draw the same number of observations from the subsets of the four labels in \emph{equal} sampling, i.e.\ 25 observations each in the scenario of $n=100$. Using \emph{random} sampling, we make a simple random selection from the total number of all possible training examples and with \emph{stratified} sampling we ensure that the label distribution of the entire training set is replicated as accurately as possible even in smaller samples. We create 5 datasets for each combination of sampling strategy (Equal, Random, Stratified) and number of shots ($n=10, 100, 250$), where we do not consider to random sample only 10 observations in order to ensure a minimum of one observation per label in all training sets. Table~\ref{tab:ukraine_n} provides the label distributions of the training and test dataset and the corresponding expected numbers under random sampling with $n=100, 250$.

\subsection{Experimental Setup}
For the comparison of our PETapter model to PET and PEFT with a linear layer as classification head, we use the XLM-RoBERTa Large model due to the German dataset. In addition to the different architectures (cf.~Section~\ref{sec:lab_exp}), we compare the effect of different sampling strategies and the number of training observations $n=10, 100, 250$. We again repeat each experiment five times. Thus, a single cell in Table~\ref{tab:f1_ukraine} corresponds to 25 experiments (5 repetitions $\times$ 5 datasets). In Appendix~\ref{sec:appendix_ukraine}, further information on all parameters as well as the used patterns is given.

\begin{table}[t]
\small
    \centering
    \begin{tabular}{c|rrrr|r}
        \toprule 
         Label & \multicolumn{4}{c|}{Train} & Test\\
         & 10$^*$ & 100$^*$ & 250$^*$ & All & \\
         \hline
         argumentagainst & 1.2 & 11.8 & 29.4 & 92 & 118 \\
         argumentfor & 1.9 & 19.4 & 48.6 & 152 & 162 \\
         claimagainst & 2.4 & 23.5 & 58.8 & 184 & 248 \\
         claimfor & 4.6 & 45.2 & 113.2 & 354 & 456\\
         \bottomrule
    \end{tabular}
    \caption{Label distribution in the train and test data split of the Ukraine dataset. $^*$Expected distribution for random selection.
    }
    \label{tab:ukraine_n}
\end{table}

\begin{table*}[t]
    \centering
    \small
    \begin{tabular}{cc|ccc|c|ccc}
    \toprule 
    && \multicolumn{3}{c|}{\underline{PETapter}} && \multicolumn{3}{c}{\underline{Linear Layer}}\\
    $n$ & Sampling & (IA)$^3$ & LoRA & Pfeiffer & PET & (IA)$^3$ & LoRA & Pfeiffer\\
    \hline
  10 & Equal & 0.28 $\pm$.046 & 0.31 $\pm$.043 & \bf 0.33 $\pm$.057 & 0.33 $\pm$.080 & 0.14 $\pm$.039 & 0.13 $\pm$.042 & 0.15 $\pm$.041 \\
  10 & Stratified & 0.19 $\pm$.023 & 0.27 $\pm$.039 & 0.33 $\pm$.027 & \bf 0.40 $\pm$.055 & 0.16 $\pm$.000 & 0.16 $\pm$.001 & 0.17 $\pm$.021 \\ 
  \hline
  100 & Equal & 0.49 $\pm$.023 & 0.57 $\pm$.020 & 0.57 $\pm$.028 & \bf 0.59 $\pm$.027 & 0.23 $\pm$.041 & 0.26 $\pm$.029 & 0.29 $\pm$.030 \\ 
  100 & Random & 0.41 $\pm$.036 & \bf 0.56 $\pm$.036 & 0.55 $\pm$.036 & 0.56 $\pm$.053 & 0.16 $\pm$.000 & 0.20 $\pm$.041 & 0.26 $\pm$.037 \\ 
  100 & Stratified & 0.40 $\pm$.029 & 0.58 $\pm$.042 & 0.57 $\pm$.035 & \bf 0.59 $\pm$.054 & 0.16 $\pm$.000 & 0.20 $\pm$.030 & 0.26 $\pm$.035 \\ 
  \hline
  250 & Equal & 0.57 $\pm$.015 & 0.67 $\pm$.014 & 0.68 $\pm$.018 & \bf 0.70 $\pm$.025 & 0.28 $\pm$.027 & 0.46 $\pm$.050 & 0.49 $\pm$.075 \\ 
  250 & Random & 0.50 $\pm$.031 & \bf 0.67 $\pm$.021 & 0.67 $\pm$.024 & 0.67 $\pm$.109 & 0.16 $\pm$.000 & 0.38 $\pm$.031 & 0.45 $\pm$.086 \\ 
  250 & Stratified & 0.48 $\pm$.036 & 0.67 $\pm$.019 & \bf 0.67 $\pm$.018 & 0.67 $\pm$.109 & 0.16 $\pm$.003 & 0.37 $\pm$.040 & 0.46 $\pm$.082 \\ 
  \bottomrule
    \end{tabular}
    \caption{Mean macro-F1 scores ($\pm$ standard deviation) of the experiments in the real-world study (Ukraine).}
    \label{tab:f1_ukraine}
\end{table*}

\begin{table*}[t]
    \centering
    \small
    \begin{tabular}{cc|ccc|ccc}
    \toprule
    && \multicolumn{3}{c|}{\underline{Equal Sampling}} & \multicolumn{3}{c}{\underline{Stratified Sampling}}\\
    & Label & PETapter & PET & Lin.\ Layer & PETapter & PET & Lin.\ Layer\\
    \hline
  \multirow{4}{*}{\rotatebox[origin=c]{90}{Precision}} & argumentagainst & 0.50 $\pm$.031 & 0.51 $\pm$.040 & 0.34 $\pm$.068 & \bf 0.54 $\pm$.037 & 0.54 $\pm$.122 & 0.39 $\pm$.092 \\ 
  & argumentfor & 0.58 $\pm$.045 & 0.63 $\pm$.064 & 0.45 $\pm$.075 & 0.64 $\pm$.032 & \bf 0.66 $\pm$.145 & 0.47 $\pm$.135 \\ 
  & claimagainst & 0.70 $\pm$.033 & \bf 0.71 $\pm$.045 & 0.50 $\pm$.078 & 0.68 $\pm$.036 & 0.66 $\pm$.143 & 0.49 $\pm$.065 \\ 
  & claimfor & 0.87 $\pm$.022 & \bf 0.90 $\pm$.023 & 0.69 $\pm$.076 & 0.84 $\pm$.022 & 0.83 $\pm$.080 & 0.65 $\pm$.064 \\ 
  \hline
  \multirow{4}{*}{\rotatebox[origin=c]{90}{Recall}} & argumentagainst & 0.75 $\pm$.043 & \bf 0.81 $\pm$.061 & 0.46 $\pm$.113 & 0.53 $\pm$.064 & 0.57 $\pm$.140 & 0.16 $\pm$.098 \\ 
  & argumentfor & 0.66 $\pm$.048 & \bf 0.70 $\pm$.047 & 0.65 $\pm$.060 & 0.63 $\pm$.053 & 0.61 $\pm$.138 & 0.42 $\pm$.147 \\ 
  & claimagainst & 0.75 $\pm$.026 & 0.76 $\pm$.051 & 0.51 $\pm$.090 & \bf 0.78 $\pm$.028 & 0.75 $\pm$.161 & 0.54 $\pm$.121 \\ 
  & claimfor & 0.67 $\pm$.034 & 0.67 $\pm$.047 & 0.48 $\pm$.121 & 0.78 $\pm$.022 & \bf 0.80 $\pm$.051 & 0.74 $\pm$.051 \\
  \hline
  \multirow{4}{*}{\rotatebox[origin=c]{90}{Macro-F1}} & argumentagainst & 0.60 $\pm$.029 & \bf 0.62 $\pm$.031 & 0.38 $\pm$.082 & 0.53 $\pm$.045 & 0.55 $\pm$.123 & 0.22 $\pm$.100 \\ 
  & argumentfor & 0.62 $\pm$.026 & \bf 0.66 $\pm$.040 & 0.53 $\pm$.061 & 0.63 $\pm$.024 & 0.63 $\pm$.134 & 0.44 $\pm$.138 \\ 
  & claimagainst & 0.72 $\pm$.019 & \bf 0.73 $\pm$.025 & 0.50 $\pm$.072 & 0.72 $\pm$.028 & 0.70 $\pm$.148 & 0.51 $\pm$.084 \\ 
  & claimfor & 0.76 $\pm$.018 & 0.77 $\pm$.031 & 0.56 $\pm$.105 & \bf 0.81 $\pm$.012 & 0.81 $\pm$.040 & 0.69 $\pm$.049 \\
  \bottomrule
    \end{tabular}
    \caption{Mean precision, recall, and macro-F1 scores per label (each $\pm$ standard deviation) of the experiments in the real-world study (Ukraine). We consider the Pfeiffer adapter as the PEFT method of PETapter and $n=250$. We omit the results using random sampling as they are quite similar to those of the stratified datasets, cf.~Table~\ref{tab:precall_ukraine_random}.}
    \label{tab:precall_ukraine}
\end{table*}

\begin{table*}[t]
    \centering
    \small
    \begin{tabular}{cc|ccc|ccc}
    \toprule 
    && \multicolumn{3}{c|}{\underline{No Pattern}} & \multicolumn{3}{c}{\underline{Pattern}}\\
    $n$ & Sampling & Alpha & Normal & Shuffle & Alpha & Normal & Shuffle\\
    \hline
  10 & Equal & 0.22 $\pm$.039 & 0.25 $\pm$.040 & 0.22 $\pm$.039 & 0.23 $\pm$.039 & 0.31 $\pm$.043 & 0.28 $\pm$.043 \\ 
  10 & Stratified & 0.20 $\pm$.025 & 0.22 $\pm$.031 & 0.22 $\pm$.033 & 0.23 $\pm$.067 & 0.27 $\pm$.039 & 0.27 $\pm$.043 \\ 
  \hline
  100 & Equal & 0.47 $\pm$.026 & 0.43 $\pm$.041 & 0.41 $\pm$.037 & 0.57 $\pm$.041 & 0.57 $\pm$.020 & 0.56 $\pm$.033 \\ 
  100 & Random & 0.43 $\pm$.046 & 0.39 $\pm$.040 & 0.39 $\pm$.043 & 0.53 $\pm$.035 & 0.56 $\pm$.036 & 0.54 $\pm$.044 \\ 
  100 & Stratified & 0.40 $\pm$.027 & 0.38 $\pm$.035 & 0.37 $\pm$.027 & 0.52 $\pm$.051 & 0.58 $\pm$.042 & 0.54 $\pm$.048 \\ 
  \hline
  250 & Equal & 0.62 $\pm$.022 & 0.61 $\pm$.024 & 0.60 $\pm$.022 & 0.67 $\pm$.021 & 0.67 $\pm$.014 & 0.68 $\pm$.019 \\ 
  250 & Random & 0.58 $\pm$.035 & 0.57 $\pm$.054 & 0.56 $\pm$.054 & 0.65 $\pm$.029 & 0.67 $\pm$.021 & 0.66 $\pm$.020 \\ 
  250 & Stratified & 0.60 $\pm$.025 & 0.59 $\pm$.030 & 0.58 $\pm$.032 & 0.65 $\pm$.019 & 0.67 $\pm$.019 & 0.66 $\pm$.017 \\ 
  \bottomrule
    \end{tabular}
    \caption{Mean macro-F1 scores ($\pm$ standard deviation) of the PVP-experiments in the real-world study (Ukraine) using LoRA as PEFT method and PETapter as classification head.}
    \label{tab:f1_pvp_ukraine}
\end{table*}

\subsection{Results}
\label{sec:results_ukraine}
Table~\ref{tab:f1_ukraine} shows the macro-F1 scores from the real-world study. According to this, there is no significant difference between the performance of PET and PETapter. The scores using the linear layer, on the other hand, are significantly worse in all scenarios and combinations. The classification task appears to be so hard that for $n=10$ neither any scenario nor any model could achieve meaningful performance. For $n=100, 250$, it can be seen that PET benefits greatly from the use of a balanced training set (equal sampling); this can be seen in a strong reduction of uncertainty measured by the standard deviation. In principle, however, PETapter produces consistently more reliable performances in the sense that the standard deviation of the scores is consistently lower for all settings. It can be seen that even for $n=100$ random and stratified sampling lead to similar results.

In Table~\ref{tab:precall_ukraine}, we also present label-specific performance scores (precision, recall, macro-F1) for each of the four classes in the dataset. Here, we focus on the presentation of the results for $n=250$ and the Pfeiffer adapters as PEFT method. The results of the random sampling are shown in Table~\ref{tab:precall_ukraine_random} in the Appendix. It is of little surprise that the rarest class \emph{argumentagainst} generates the worst precision, recall, and macro-F1 scores in the stratified sampling. In the case of equal sampling, this remains true only for the precision; the recall can be increased through the oversampling for PET from previously 51\% to the then highest value of the four classes of 81\%. The fact that an equal sampling in the training set leads to higher overall macro-F1 scores and lower uncertainty than a stratified sampling can be explained by the fact that the corresponding recall values of otherwise rarely occurring classes can be increased in this way.

In general, PETapter provides the best overall performance in this real-world study. While it yields a similar level of performance, it produces more reliable performance values overall. Thus, PETapter makes the idea of PET easily accessible in PEFT settings without loss of performance. While PET on our system\footnote{48\,GB NVIDIA RTX 6000 Ada, Intel Xeon W7-3445 20$\times$2.6\,GHz, 256\,GB ECC  DDR5-4800 RAM} processes 7 observations per second (obs/s) during training and 8~obs/s during testing, PETapter processes 25~obs/s during training and 51~obs/s ($(\text{IA})^3$/LoRA) or 42~obs/s (Pfeiffer) during testing. This shows a meaningful speed-up of PETapter compared to PET for the training as well as the inference phase.

\subsection{PVP-Experiments}
A major criticism of PET-like models is the need for manual generation of patterns and verbalizers. To address this, we tested the use of automated PVPs (\emph{No Pattern}, \emph{Alpha} verbalizer) and badly chosen verbalizers (\emph{Shuffle}). For reasons of complexity, we limit the experiment to the combination of LoRA and PETapter. As a result from Table~\ref{tab:f1_pvp_ukraine}, it can be concluded that the pattern should at best be chosen manually, as there are notable differences between the performances in all scenarios. However, the choice of verbalizer in our task has hardly any influence on the performance already for $n=100$. In fact, for \emph{No Pattern}, \emph{Alpha} mostly performs better than the manually selected verbalizers, which may be because \emph{Alpha} consists of only one token each, while the other two scenarios require two tokens per verbalizer. Moreover, \emph{Shuffle} does not result in any noteworthy differences in label-specific performance values compared to \emph{Normal} (table not included due to space constraints).

As an extension, Table~\ref{tab:f1_majority} indicates that a combination of the five repetitions using a majority vote leads to superior and more stable results. In particular in the scenario of limited human resources, this offers the possibility of boosting the performance of automatically selected PVPs by simply stacking independent repetitions.

\begin{table}[t]
    \centering
    \small
    \begin{tabular}{ccc|cc}
    \toprule
Data & $n$ & PVP & Mean & Majority \\ 
  \hline
AG &  10 & Manual & 0.71 $\pm$.080 & 0.71 $\pm$.071 \\ 
AG & 100 & Manual & 0.87 $\pm$.010 & 0.87 $\pm$.009 \\ 
Yahoo &  10 & Manual & 0.30 $\pm$.044 & 0.32 $\pm$.034 \\ 
Yahoo & 100 & Manual & 0.66 $\pm$.013 & 0.66 $\pm$.012 \\ 
Yelp &  10 & Manual & 0.45 $\pm$.055 & 0.45 $\pm$.053 \\ 
Yelp & 100 & Manual & 0.61 $\pm$.013 & 0.62 $\pm$.009 \\ 
Ukraine &  10 & Manual & 0.27 $\pm$.039 & 0.27 $\pm$.029 \\
Ukraine & 100 & Manual & 0.58 $\pm$.042 & 0.59 $\pm$.025 \\
Ukraine & 250 & Manual & 0.67 $\pm$.019 & 0.69 $\pm$.018 \\ 
Ukraine &  10 & Autom. & 0.20 $\pm$.025 & 0.20 $\pm$.021 \\ 
Ukraine & 100 & Autom. & 0.40 $\pm$.027 & 0.41 $\pm$.018 \\ 
Ukraine & 250 & Autom. & 0.60 $\pm$.025 & 0.62 $\pm$.025 \\ 
  \bottomrule
    \end{tabular}
    \caption{Mean (majority) macro-F1 scores ($\pm$ standard deviation) using LoRA and PETapter. For Ukraine, we present the results using the \emph{Stratified} sampling. \emph{Manual} PVP means \emph{Prompt} pattern for AG, Yahoo, and Yelp; for Ukraine, it represents the combination of \emph{Pattern} using the \emph{Normal} verbalizers, \emph{Autom.} represents the combination of \emph{No Pattern} and \emph{Alpha}.}
    \label{tab:f1_majority}
\end{table}

\section{Discussion}
\label{sec:discussion}
The results show that PET benefits greatly from equal sampling with unbalanced data. We were not able to achieve this gain to the same extent using PETapter. Therefore, even though equal sampling is not a realistic scenario in (few-shot) real-world settings, it could be promising to develop PEFT methods in such a way that they benefit from balanced data as much as PET. Moreover, as a follow-up study, we want to investigate to what extent or at what level of unbalancedness it is worth simply omitting observations from a stratified sampling in order to obtain a more balanced training set.

Following the mathematical definition of PETapter (cf.~Equation~\ref{eq:score}), it is recommended to always use the same number of verbalizer tokens for each class. We also recommend using as simple patterns as possible. Initial experiments have shown that PET in particular leads to poor results with small $n$ if the pattern is too complex. PETapter was slightly more robust against the choice of pattern in these experiments. In addition, although grammatical correctness of the pattern is desirable, it is not essential to achieve satisfactory results.

For the activation function, we decided to use GELU in combination with LayerNorm (cf.~Figure~\ref{fig:petapter}). We will further investigate this decision in future studies by examining the influence of using alternative activation functions.

\section{Conclusion}
\label{sec:conclusion} 
We show that our novel method PETapter combines the advantages of few-shot learning and parameter-efficient fine-tuning (PEFT). PETapter combines PEFT methods with PET-style classification heads. In this way, it makes the idea of PET easily accessible in PEFT settings. As a result, it achieves PET performance and increased reliability of the results while offering all the advantages of PEFT. It can be trained faster with higher parameter efficiency and without catastrophic forgetting. Furthermore, it offers a modularity that makes it easier to share models because the newly learned parameters require, e.g., only 16\,MB compared to 2.1\,GB of disk space. Due to the implementation in the Adapters \cite{poth-etal-2023-adapters} framework, it allows easy execution and combination with existing/new PEFT methods like QLoRA \cite{dettmers2023qlora}, mixture of experts models \cite{anonymous2023pushing}, or others (cf.~Section~\ref{sec:peft}). Our method makes high-performing text classification --- which is beneficial to many research domains where labeled text is a valuable resource --- feasible for more researchers, by lowering the technological boundaries to accessing language models.  


\section*{Acknowledgements} 
This work was funded by the Bundesministerium für Bildung und Forschung (BMBF) as part of the project \emph{FLACA: Few-shot learning for automated content analysis in communication science} (project no.\ 16DKWN064B) and by the Deutsche Forschungsgemeinschaft (DFG) as part of the project \emph{FAME: A framework for argument mining and evaluation} (project no.\ 406289255). In addition, we acknowledge the computation resources facilitated by seed funding from the Research Center Trustworthy Data Science and Security as part of the project \emph{Trustworthy performance evaluation of large language models}.

\section*{Limitations}
As mentioned in Sections~\ref{sec:discussion} and~\ref{sec:conclusion}, not all promising models and combinations of model ingredients can be evaluated. Therefore, it is expected that better performance scores can be achieved by optimizing the combination of all mentioned ingredients, cf.~Section~\ref{sec:related_work}. Instead, we show that the PETapter idea is a promising method overall, if the underlying task is possible to formulate as classification tasks. In addition, our analyses are limited to four data sets (also for reasons of sustainable NLP), which are restricted to the languages English and German. In return, we pay a lot of attention to a reliable evaluation through 5 repetitions $\times$ 5 sampled datasets for each of the considered scenarios in Sections~\ref{sec:lab} and~\ref{sec:ukraine}.


\bibliography{anthology,custom}

\appendix

\section{Details of Benchmark Study}
\label{sec:appendix_ag_yelp_yahoo}
For the benchmark study, we compare the use of \enquote{roberta-base} and \enquote{roberta-large} from HuggingFace's Transformers \cite{wolf-etal-2020-transformers}. For PET (\url{https://github.com/timoschick/pet}), we use the two parameters \texttt{pet\_num\_train\_epochs=10} and \texttt{pet\_per\_gpu\_train\_batch\_size=1} that differ from the default. For all experiments using the package Adapter \cite{poth-etal-2023-adapters}, we use the parameters
\begin{compactitem}
 \item \texttt{c\_rate=16} (Pfeiffer),
 \item \texttt{r=8} (LoRA),
 \item \texttt{alpha=16} (LoRA),
 \item \texttt{learning\_rate=5.0e-5},
 \item \texttt{max\_epochs=30},
 \item \texttt{per\_device\_train\_batch\_size=2}
\end{compactitem}
and alternate \texttt{arch} with \texttt{ia3}, \texttt{lora}, and \texttt{pfeiffer}. The patterns in the following subsections are motivated by the results of the study of \citet{schick-schutze-2022-true}. Due to the limited input length of PLMs, potential truncations of the input elements in the pattern are indicated with *, potentially as group within \{\} brackets.

\subsection{AG's News}
This English language dataset is available at \url{https://huggingface.co/datasets/ag_news} \cite{datasets} and consists of 120 thousand training and 7.6 thousand test observations.
\begin{description}
    \item[Prompt Pattern] [MASK] News: [text]*
    \item[Q\&A Pattern] [text]* [SEP] Question: What is the topic of this article? Answer: [MASK].
    \item[Verbalizers] \phantom{A}
    \begin{description}
        \item[World] World
        \item[Sports] Sports
        \item[Business] Business
        \item[Sci/Tech] Tech
    \end{description}
\end{description}

\subsection{Yahoo Questions}
This English language dataset is available at \url{https://huggingface.co/datasets/yahoo_answers_topics} \cite{datasets} and consists of 1.4 million training and 60 thousand test observations.
\begin{description}
    \item[Prompt Pattern] [MASK] Question: \{[ques\-tion\-\_title] [ques\-tion\-\_content] [best\-\_answer]\}*
    \item[Q\&A Pattern] \{[question\-\_title] [question\-\_con\-tent] [best\-\_answer]\}* [SEP] Question: What is the topic of this question? Answer: [MASK].
    \item[Verbalizers] \phantom{A}
    \begin{description}
        \item[Society \& Culture] Society
        \item[Science \& Mathematics] Science
        \item[Health] Health
        \item[Education \& Reference] Education
        \item[Computers \& Internet] Computer
        \item[Sports] Sports
        \item[Business \& Finance] Business
        \item[Entertainment \& Music] Entertainment
        \item[Family \& Relationships] Relationship
        \item[Politics \& Government] Politics
    \end{description}
\end{description}

\begin{table*}[t]
\small
    \centering
    \begin{tabular}{cc|ccc|ccc}
    \toprule
    && \multicolumn{3}{c|}{\underline{Random Sampling}} & \multicolumn{3}{c}{\underline{Stratified Sampling}}\\
    & Label & PETapter & PET & Lin.\ Layer & PETapter & PET & Lin.\ Layer\\
    \hline
  \multirow{4}{*}{\rotatebox[origin=c]{90}{Precision}} & argumentagainst & 0.57 $\pm$.042 & 0.57 $\pm$.126 & 0.37 $\pm$.228 & 0.54 $\pm$.037 & 0.54 $\pm$.122 & 0.39 $\pm$.092 \\ 
  & argumentfor & 0.64 $\pm$.050 & 0.66 $\pm$.141 & 0.45 $\pm$.117 & 0.64 $\pm$.032 & 0.66 $\pm$.145 & 0.47 $\pm$.135 \\ 
  & claimagainst & 0.68 $\pm$.030 & 0.66 $\pm$.141 & 0.52 $\pm$.063 & 0.68 $\pm$.036 & 0.66 $\pm$.143 & 0.49 $\pm$.065 \\ 
  & claimfor & 0.81 $\pm$.027 & 0.81 $\pm$.077 & 0.65 $\pm$.065 & 0.84 $\pm$.022 & 0.83 $\pm$.080 & 0.65 $\pm$.064 \\ 
  \hline
  \multirow{4}{*}{\rotatebox[origin=c]{90}{Recall}} & argumentagainst & 0.51 $\pm$.059 & 0.53 $\pm$.147 & 0.12 $\pm$.093 & 0.53 $\pm$.064 & 0.57 $\pm$.140 & 0.16 $\pm$.098 \\ 
  & argumentfor & 0.61 $\pm$.056 & 0.61 $\pm$.140 & 0.42 $\pm$.169 & 0.63 $\pm$.053 & 0.61 $\pm$.138 & 0.42 $\pm$.147 \\ 
  & claimagainst & 0.75 $\pm$.039 & 0.74 $\pm$.157 & 0.54 $\pm$.175 & 0.78 $\pm$.028 & 0.75 $\pm$.161 & 0.54 $\pm$.121 \\ 
  & claimfor & 0.80 $\pm$.029 & 0.82 $\pm$.047 & 0.76 $\pm$.054 & 0.78 $\pm$.022 & 0.80 $\pm$.051 & 0.74 $\pm$.051 \\
  \hline
  \multirow{4}{*}{\rotatebox[origin=c]{90}{Macro-F1}} & argumentagainst & 0.53 $\pm$.037 & 0.55 $\pm$.128 & 0.16 $\pm$.109 & 0.53 $\pm$.045 & 0.55 $\pm$.123 & 0.22 $\pm$.100 \\ 
  & argumentfor & 0.62 $\pm$.046 & 0.63 $\pm$.134 & 0.43 $\pm$.142 & 0.63 $\pm$.024 & 0.63 $\pm$.134 & 0.44 $\pm$.138 \\ 
  & claimagainst & 0.71 $\pm$.025 & 0.70 $\pm$.146 & 0.52 $\pm$.124 & 0.72 $\pm$.028 & 0.70 $\pm$.148 & 0.51 $\pm$.084 \\ 
  & claimfor & 0.81 $\pm$.016 & 0.81 $\pm$.039 & 0.70 $\pm$.036 & 0.81 $\pm$.012 & 0.81 $\pm$.040 & 0.69 $\pm$.049 \\
  \bottomrule
    \end{tabular}
    \caption{Mean precision, recall, and macro-F1 scores per label (each $\pm$ standard deviation) of the experiments in the real-world study (Ukraine). We consider the Pfeiffer adapter as the PEFT method of PETapter and $n=250$. In addition to the results from Table~\ref{tab:precall_ukraine}, we present the results of random sampling in comparison to stratified sampling.}
    \label{tab:precall_ukraine_random}
\end{table*}

\subsection{Yelp Full}
This English language dataset is available at \url{https://huggingface.co/datasets/yelp_review_full} \cite{datasets} and consists of 650 thousand training and 50 thousand test observations.
\begin{description}
    \item[Prompt Pattern] [text]* [SEP] All in all, it was [MASK].
    \item[Q\&A Pattern] [text]* [SEP] Question: What does the customer think of this restaurant? Answer: [MASK].
    \item[Verbalizers] \phantom{A}
    \begin{description}
        \item[1 star] terrible
        \item[2 stars] bad
        \item[3 stars] okay
        \item[4 stars] good
        \item[5 stars] great
    \end{description}
\end{description}

\section{Details of the Real-World Study}
\label{sec:appendix_ukraine}

The dataset will be made available to researchers after acceptance of the paper. Since this dataset is in German language, we use \enquote{xlm-roberta-large} from HuggingFace's Transformers \cite{wolf-etal-2020-transformers}. For PET (\url{https://github.com/timoschick/pet}), we use the two parameters \texttt{pet\_num\_train\_epochs=10} and \texttt{pet\_per\_gpu\_train\_batch\_size=1} (for a fair training time comparison in Table~\ref{tab:times_clinic}, we use \texttt{pet\_per\_gpu\_train\_batch\_size=2}) that differ from the default. For all experiments using the package Adapter \cite{poth-etal-2023-adapters}, we use the parameters
\begin{compactitem}
 \item \texttt{c\_rate=16} (Pfeiffer),
 \item \texttt{r=8} (LoRA),
 \item \texttt{alpha=16} (LoRA),
 \item \texttt{learning\_rate=5.0e-5},
 \item \texttt{max\_epochs=30},
 \item \texttt{per\_device\_train\_batch\_size=2}
\end{compactitem}
and alternate \texttt{arch} with \texttt{ia3}, \texttt{lora}, and \texttt{pfeiffer}.

Due to the limited input length of PLMs, potential truncations of the input elements in the pattern are indicated with *, potentially as group within \{\} brackets. We place the target sentence at the beginning to ensure that it is never truncated due to restrictions regarding the model's input length of 512 tokens. Experiments with target\_sentence in the middle returned slightly worse results. In the same way, experiments with an equivalent German pattern yield slightly worse results.

\begin{description}
    \item[Pattern] This sentence contains [MASK] [MASK] arms deliveries to Ukraine: \{[target\_sentence] [SEP] [context\_before] [SEP] [context\_after]\}*
    \item[No Pattern] [MASK] [MASK]: \{[target\_sentence] [SEP] [context\_before] [SEP] [context\_after]\}*
    \item[Alpha Verbalizers] \phantom{A}\\
    The alphanumeric verbalizers only consist of one token each, so that in this case the used pattern is reduced to one [MASK] token only.
    \begin{description}
        \item[argumentagainst] a
        \item[argumentfor] b
        \item[claimagainst] c
        \item[claimfor] d
        \item[nostance] e
    \end{description}
    \item[Normal Verbalizers] \phantom{A}
    \begin{description}
        \item[argumentagainst] argument against
        \item[argumentfor] argument for
        \item[claimagainst] claim against
        \item[claimfor] claim for
        \item[nostance] nothing regarding
    \end{description}
    \item[Shuffle Verbalizers] \phantom{A}
    \begin{description}
        \item[argumentagainst] claim for
        \item[argumentfor] claim against
        \item[claimagainst] nothing regarding
        \item[claimfor] argument against
        \item[nostance] argument for
    \end{description}
\end{description}

\section{RAFT Benchmark}
\label{sec:appendix_raft}
The RAFT Leaderboard \cite[\url{https://raft.elicit.org/}]{alex2021raft} has been under maintenance for some time now. We made a submission a while ago, but unfortunately still got no score. If the scores are published during the review process, we will include a table with comparative values for T-Few \cite{liu2022ia3}, Setfit \cite{tunstall2022efficient}, PET \cite{schick-schutze-2021-exploiting} and the human baseline at this point.

The 11 datasets of RAFT are in English language and available at \url{https://huggingface.co/datasets/ought/raft}. We use the model \enquote{microsoft/deberta-v2-xxlarge} from HuggingFace's Transformers \cite{wolf-etal-2020-transformers} and for all 11 datasets the parameters
\begin{compactitem}
\item \texttt{arch=lora},
\item \texttt{r=8},
\item \texttt{alpha=16},
\item \texttt{learning\_rate=5.0e-5},
\item \texttt{max\_epochs=30},
\item \texttt{per\_device\_train\_batch\_size=2},
\item \texttt{number\_of\_runs=5},
\end{compactitem}
where we select the majority vote out of the five runs as the final submission.

The patterns in the following subsections are motivated by the results of the study of \citet{schick-schutze-2022-true}. Due to the limited input length of PLMs, potential truncations of the input elements in the patterns are indicated with *, potentially as group within \{\} brackets.

\subsection{ade\_corpus\_v2}
\begin{description}
    \item[Pattern] [Sentence]* [SEP] Question: Is this sentence related to an adverse drug effect (ADE)? Answer: [MASK].
    \item[Verbalizers] \phantom{A}
    \begin{description}
        \item[not ADE-related] No
        \item[ADE-related] Yes
    \end{description}
\end{description}

\subsection{banking\_77}
For the processing of the \emph{banking\_77} dataset, some preprocessing is necessary. As there are only 50 observations in each of the training sets of the RAFT setting, but at the same time there are 77 different classes in this specific dataset, the model misses 27 of the classes in the training set. To overcome this, we augment the training data such that each of the 50 observations is combined with all 77 classes using the yes/no pattern below. As a result, the 27 classes that were previously not included in the training data now become part it each 50 times in combination with the label \emph{no}.
\begin{description}
    \item[Pattern] The following is a banking customer service query. [SEP] \{[Query] [SEP] Is [Label]\}* the correct category for this query? Answer: [MASK].
    \item[Verbalizers] \phantom{A}
    \begin{description}
        \item[No] No
        \item[Yes] Yes
    \end{description}
\end{description}

\subsection{neurips\_impact\_statement\_risks}
\begin{description}
    \item[Pattern] \{Title: [Paper title] [SEP] Statement: [Impact statement]\}* [SEP] Question: Does this impact statement mention a harmful application? Answer: [MASK].
    \item[Verbalizers] \phantom{A}
    \begin{description}
        \item[doesn't mention a harmful application] No
        \item[mentions a harmful application] Yes
    \end{description}
\end{description}

\subsection{one\_stop\_english}
\begin{description}
    \item[Pattern] [Article]* [SEP] Question: Is the level of this article 'elementary', 'intermediate' or 'advanced'? Answer: [MASK].
    \item[Verbalizers] \phantom{A}
    \begin{description}
        \item[elementary] elementary
        \item[intermediate] intermediate
        \item[advanced] advanced
    \end{description}
\end{description}

\subsection{overruling}
\begin{description}
    \item[Pattern] In law, an overruling sentence is a statement that nullifies a previous case decision as a precedent. [SEP] [Sentence]* [SEP] Question: Is this sentence overruling? Answer: [MASK].
    \item[Verbalizers] \phantom{A}
    \begin{description}
        \item[not overruling] No
        \item[overruling] Yes
    \end{description}
\end{description}

\subsection{semiconductor\_org\_types}
\begin{description}
    \item[Pattern] \{Title: [Paper title] [SEP] Organization name: [Organization name]\}* [SEP] Question: What is the category of this institution? Answer: [MASK].
    \item[Verbalizers] \phantom{A}
    \begin{description}
        \item[company] Company
        \item[research institute] Institute
        \item[university] University
    \end{description}
\end{description}

\subsection{systematic\_review\_inclusion}
\begin{description}
    \item[Pattern] \{Journal: [Journal] [SEP] Title: [Title] [SEP] Abstract: [Abstract]\}* [SEP] Question: Should this paper be included in a meta-review which includes the findings of systematic reviews on interventions designed to promote charitable donations? Answer: [MASK].
    \item[Verbalizers] \phantom{A}
    \begin{description}
        \item[not included] No
        \item[included] Yes
    \end{description}
\end{description}

\subsection{tai\_safety\_research}
\begin{description}
    \item[Pattern] Transformative AI (TAI) is defined as AI that precipitates a transition comparable to (or more significant than) the agricultural or industrial revolution [SEP] \{Journal: [Publication Title] [SEP] Title: [Title] [SEP] Abstract: [Abstract Note]\}* [SEP] Question: Is this paper a TAI safety research paper? Answer: [MASK].
    \item[Verbalizers] \phantom{A}
    \begin{description}
        \item[not TAI safety research] No
        \item[TAI safety research] Yes
    \end{description}
\end{description}

\subsection{terms\_of\_service}
\begin{description}
    \item[Pattern] The following sentence is from a Terms of Service. [SEP] [Sentence]* [SEP] Question: Is this sentence potentially unfair? Answer: [MASK].
    \item[Verbalizers] \phantom{A}
    \begin{description}
        \item[not potentially unfair] No
        \item[potentially unfair] Yes
    \end{description}
\end{description}

\subsection{tweet\_eval\_hate}
\begin{description}
    \item[Pattern] [Tweet]* [SEP] Question: Does this tweet contain hate speech against either immigrants or women? Answer: [MASK].
    \item[Verbalizers] \phantom{A}
    \begin{description}
        \item[not hate speech] No
        \item[hate speech] Yes
    \end{description}
\end{description}

\subsection{twitter\_complaints}
\begin{description}
    \item[Pattern] [Tweet text]* [SEP] Question: Does this tweet contain a complaint? Answer: [MASK].
    \item[Verbalizers] \phantom{A}
    \begin{description}
        \item[no complaint] No
        \item[complaint] Yes
    \end{description}
\end{description}

\section{GPT Comparison}
A different version of the Ukraine dataset, in which not only the four classes outlined in this paper are considered, but also \emph{irrelevant} sentences are contained, is studied by \citet{rieger2023flaca} in a comparison between full fine-tuning methods, PET and PEFT methods. The authors show that PETapter already with only 272 observations outperforms zero-shot GPT-4. The authors further observe that GPT-4 performs better on this real-world and unpublished dataset in a zero-shot manner than using an in-context learning prompt with few (up to 10) examples. Moreover, they found out that GPT-3.5 performs significantly worse than GPT-4 on this task.

\end{document}